\documentclass[letterpaper, 11 pt, conference]{ieeeconf}  

\IEEEoverridecommandlockouts                              
\usepackage{array}
\usepackage[caption=false,labelfont=sf,textfont=sf]{subfig}
\usepackage{textcomp}
\usepackage{stfloats}
\usepackage{url}
\usepackage{verbatim}
\usepackage{graphicx}
\usepackage{cite}
\usepackage[letterpaper, left=1in, right=1in, bottom=1in, top=1in]{geometry}

\usepackage{amsmath} 
\usepackage{comment}
\onecolumn
\title{\bf
FCOM: A Federated Collaborative Online Monitoring Framework
via Representation Learning \vspace{-2ex}
}

\author{Tanapol Kosolwattana$^1$, Huazheng Wang$^2$, Raed Al Kontar$^3$, Ying Lin$^1$
}
\usepackage{mathtools, nccmath}
\IEEEaftertitletext{\vspace{-1.1\baselineskip}}

\usepackage{algorithm}
\usepackage{algpseudocode}
\usepackage{graphicx}
\usepackage{booktabs}
\usepackage[compact]{titlesec}
\titlespacing{\section}{0pt}{*0}{*0}
\titlespacing{\subsection}{0pt}{*0}{*0}
\titlespacing{\subsubsection}{0pt}{*0}{*0}
\usepackage{etoolbox}

\usepackage{caption}

\newtheorem{theorem}{Theorem}
\newtheorem{lemma}{Lemma}
\newtheorem{proposition}{Proposition}

\usepackage{amsfonts}
\usepackage{mathrsfs}  
\usepackage{tabularx}
\makeatletter
\newcommand{\multiline}[1]{%
  \begin{tabularx}{\dimexpr\linewidth-\ALG@thistlm}[t]{@{}X@{}}
    #1
  \end{tabularx}
}
\makeatother
\begin{document}

\maketitle
\stepcounter{footnote}
\footnotetext{Department of Industrial Engineering, University of Houston}
\stepcounter{footnote}
\footnotetext{School of Electrical Engineering and Computer Science, Oregon State University}
\stepcounter{footnote}
\footnotetext{Department of Industrial \& Operations Engineering, University of Michigan}
\pagestyle{plain}


\begin{abstract}
Online learning has demonstrated notable potential to dynamically allocate limited resources to monitor a large population of processes, effectively balancing the exploitation of processes yielding high rewards, and the exploration of uncertain processes. However, most online learning algorithms were designed under 1) a centralized setting that requires data sharing across processes to obtain an accurate prediction or 2) a homogeneity assumption that estimates a single global model from the decentralized data. To facilitate the online learning of heterogeneous processes from the decentralized data, we propose a federated collaborative online monitoring method, which captures the latent representative models inherent in the population through representation learning and designs a novel federated collaborative UCB algorithm to estimate the representative models from sequentially observed decentralized data. The efficiency of our method is illustrated through theoretical analysis, simulation studies, and decentralized cognitive degradation monitoring in Alzheimer's disease.

\end{abstract}
\section{Introduction}
Monitoring a large population of dynamic processes within the constraints of monitoring resources poses a significant challenge across various industrial sectors, including healthcare and engineering systems \cite{doi:10.1080/24725854.2018.1470357, kosolwattana2023online}. The complexity arises from two key factors: 1) the inherent disparity between the limited monitoring resources available and the large population of processes to be monitored, and 2) the uncertain and heterogeneous dynamics in the progression of these processes. In tackling this intricate problem, online learning from bandit feedback has demonstrated notable potential \cite{fouche2019scaling, kosolwattana2023online}. These methods offer a solution by dynamically allocating limited resources, effectively balancing the exploitation of processes yielding high rewards and the exploration of uncertain processes. While existing online learning algorithms, including those cited above, show promise, they still exhibit certain limitations that hinder their effectiveness in addressing the complexities of monitoring under resource constraints.
The primary limitation of existing algorithms stems from their reliance on a centralized setting, necessitating the transmission of all data to a central server for model training. While this approach has demonstrated utility, it raises profound concerns about data privacy. In healthcare applications, such as patient health monitoring systems, this centralized model poses a substantial threat to data privacy, particularly with electronic health records (EHRs) containing sensitive patient information that ideally should remain within local organizations for privacy protection \cite{xu2021federated}. Beyond privacy, issues related to the costs of communication and data storage in a central server further underscore the imperative for a decentralized approach \cite{kontar2021internet}. Therefore, it is crucial to develop decentralized algorithms that empower the storage and analysis of monitoring data at the local level, mitigating privacy risks and reducing communication and storage costs. 

An ostensibly straightforward method to achieve decentralized monitoring is by modeling and monitoring the progression dynamic of each process independently using conventional online learning algorithms \cite{wang2019distributed}. However, this approach can be less effective due to the uneven data distribution among individual processes, where the disparity in available data for each process can lead to sub optimal model performance. \cite{penny2012approaches}.  
Given that each process features a heterogeneous dynamic that is not entirely independent, neglecting to exploit information sharing between processes may overlook opportunities to effectively span the model space for these processes' models \cite{8169076}. 

Several federated online learning algorithms have been developed in the literature to learn a global model under distributed datasets online. A notable approach in this context is federated multi-armed bandits (FMAB) \cite{shi2021federated, dubey2020differentially}, which provides a framework for units or processes to collaboratively solve a global bandit problem while retaining their information locally. In this approach, at each time step, each unit estimates the model locally and sends updates of model statistics to the central server. Subsequently, the central server sends back the aggregated statistics to enable each unit to enhance its model estimation for the next trial. This collaborative process allows the central server to develop a better selection strategy, striking a balance between exploration and exploitation. Nevertheless, these federated online learning algorithms usually focus on a homogeneous population that can be well described by a single global model. They can lead to a bias in estimating personalized models since the resulting global model only reflects the average effect, but fails to capture the between-unit variations \cite{tan2022towards}. Some existing studies adopt a personalized modeling approach, such as the mixed model \cite{shi2021federated, smith2018federated}, which uses random effects to describe personalized variation. However, the mixed model used in these studies still cannot capture the underlying group structure in a heterogeneous population \cite{di2023ppfl}. The underlying group structure represents diverse population characteristics where each group structure corresponds to one behavioral pattern or mechanism. Numerous studies have emphasized that capturing the underlying group structure could potentially enhance the accuracy of personalized modeling \cite{doi:10.1080/24725854.2018.1470357, kumar2013flexible}.

To address this problem, we consider the representation learning approach to exploit and monitor the hidden structure in a population. Representation learning is an approach that learns a representation to capture the common structure across different but related units \cite{du2020few}. It offers substantial benefits in learning individualized models from insufficient monitoring data, especially at the beginning of the monitoring period. 
However, to the best of our knowledge, there are no federated online learning algorithms that consider representation learning during unit modeling. The technical challenge is that achieving a good estimation of the latent group structure requires substantial communication between the central servers and units. Thus, a novel communication strategy is necessitated to balance the tradeoff between the algorithm performance (i.e., cumulative regret) and the communication cost. Additionally, existing federated online learning algorithms make decisions locally and independently for each unit. Instead, the online monitoring problem requires decision-making under global constraint, i.e., how to allocate limited monitoring resources, which differs from existing decentralized online learning. 

To mitigate these gaps, we propose a federated collaborative online monitoring framework. The proposed framework aims to allocate limited monitoring resources to a heterogeneous population in real-time, using federated online representation learning. It consists of three main phases: the representation learning-based modeling, the decentralized parameter estimation and communication, and the monitoring strategy design. In the first phase, we leverage a representation learning \cite{8169076} approach to model the heterogeneous population, which assumes the existence of a shared representation that captures the underlying group structure among units. The second phase introduces a novel federated collaborative online monitoring (FCOM) algorithm to estimate the shared representation from distributed and sequentially arrived data. In the third phase, a novel upper confidence bound (UCB)-based score is developed to real-time assess the uncertainty of model estimation from the FCOM algorithm and inform the optimal monitoring resource allocation that balances the exploitation of high-reward units and exploration of uncertain ones. We rigorously prove sublinear regret and communication cost upper bounds of the proposed FCOM algorithm. The efficiency of the proposed method is validated by a simulation study and an empirical study on online cognitive degradation monitoring for Alzheimer's disease (AD).

The contributions of this paper can be summarized as:
\begin{itemize}
    \item We propose a novel Federated Online Collaborative Monitoring (FCOM) algorithm, which enables decentralized online learning of the latent group structure from distributed data. It allows each unit to update the parameters locally and transmit the statistics that do not contain the true observed data to the central server.
    \item We develop an event-triggered communication strategy, allowing units to communicate or synchronize with the central server only when they have gathered sufficient new information. This helps lower communication costs while maintaining a low regret.
    \item Through regret analysis, we demonstrate that the proposed algorithm yields a reduced upper regret bound compared to the benchmark models when the latent group structure is low-rank.
    \item We demonstrate the effectiveness of the proposed method through simulation studies and an empirical study of adaptive cognitive monitoring in Alzheimer's disease. The results indicate that our method, under a federated setting, achieves comparable performance with benchmark models while preserving the data within local units.
\end{itemize}

\section{Related work} \label{Related_work}
\subsection{Online monitoring}
Online monitoring can be formulated as a sequential decision-making problem where the decision of which units need to be monitored is made in each trial/cycle. Due to the unknown dynamics in process progression, reinforcement learning methods have been recently used to solve the online monitoring problem. The selective sensing method developed in \cite{doi:10.1080/24725854.2018.1470357} 
incorporates Markov process and representation learning to model personalized process progression for a population of units and builds an optimization model to allocate resources to units that provide the highest monitoring rewards. However, it cannot balance the exploitation-exploration trade-off in resource allocation since it does not consider the uncertainty in model estimation. The collaborative learning-based UCB (CLUCB) algorithm introduced in \cite{kosolwattana2023online} mitigates this limitation by designing a novel UCB-based exploration method to assess model estimation uncertainty and find an optimal monitoring strategy under the exploitation-exploration trade-off. However, these methods, as mentioned earlier, consider the online monitoring problem in a centralized setting, meaning all units’ observations of reward and historical data are shared and exposed in the central server, which may lead to the risk of privacy leakage and high computational costs.

\subsection{Federated online learning}
In federated online learning, each unit collects the observations sequentially instead of observing the whole training data at once, updates the local model, and transmits its local updates to the central server when new observations are available, which broadcasts a global aggregated model to all units at the end of the time step \cite{mitra2021online}. In \cite{wu2023decentralized}, the unit estimates an individualized model based on its neighbor-units' models in which their learned aggregation weights are updated at each time step. Since it only exploits the information from neighboring units, it does not reflect insights about the population group structure, which requires the units to learn across groups. The federated multi-armed bandit (FMAB) is designed under a decentralized setting to preserve data privacy \cite{shi2021federated,zhu2021federated}. Distributed Linear UCB or DisLinUCB, in short, proposed in \cite{wang2019distributed} considers a star-shaped communication between units and a central server to update globally shared parameters under synchronous setting, but they assume all units are homogeneous. Some existing algorithms consider an additive model or a mixed model approach to account for both individual models and globally shared models \cite{shi2021federated, smith2018federated}. In Sync-LinUCB \cite{pmlr-v151-li22e}, the unknown parameter for each unit consists of a globally shared component, known as a fixed effect, and individualized local component, known as a random effect. The random effect can be estimated from local data and the fixed effect needs to be estimated collaboratively in the central server. However, none of these existing algorithms can capture the inherent group structures of heterogeneous populations.

\section{method} \label{Method}
\subsection{Problem Statement}

Consider a monitoring system featuring $N$ units along with a central server responsible for allocating the monitoring resources over units and coordinating communication between itself and the units. Each unit's monitoring reward is a dynamic process associated with a time-varying feature vector $x_{it} \in \mathbb{R}^{p}$. If the unit is monitored, a corresponding monitoring reward can be observed from the environment, which is predicted by the feature vector using a personalized linear reward function $y_{it} = f_{i}(x_{it}) + \epsilon_i$. $ \epsilon_i$ represents a random noise that follows a normal distribution. This reward function needs to satisfy two general assumptions, which are non-decreasing monotonicity and bounded smoothness \cite{qin2014contextual}. 
The personalized reward functions of $N$ units are assumed to be related, meaning there exists a shared representation that captures the latent structure across the unit population. Exploiting this common representation can potentially improve the quality of each unit's model prediction, which leads to better monitoring decisions. 

The primary objective of the monitoring system is to select a set $\mathscr{A}_{t}$ in each trial $t$, comprising $M$ units to be monitored, that results in the maximal monitoring reward. The total monitoring reward can be represented as:
\begin{equation}
     r_{\mathscr{A}_{t}} = \sum_{i=1}^N a_{it}y_{it}\label{eq:1}
\end{equation}
where a decision variable $a_{it} = 1$ if a unit $i \in \mathscr{A}_{t}$ or 0 otherwise and $\sum_{i=1}^N a_{it} = M$. Ideally, with the knowledge of reward functions, the optimal strategy of the central server is to choose the top $M$ units with maximal rewards in each trial, denoted as $a_{it}^*$. However, $f_i$ is unknown, and the central server needs to update its selection strategy through the learning of models from observation $(x_{it}, y_{it})$ in each trial $t$. Thus, the goal of the central server is equivalent to minimizing the cumulative regret $R(T)$ which is defined as the difference between the central server's total reward and the total reward of the optimal strategy shown as follows.
\begin{equation}
     R(T) = \sum_{t=1}^T R_t = \sum_{t=1}^T \sum_{i=1}^N a_{it}^*y_{it} - a_{it}y_{it}\label{eq:2}
\end{equation}

Additionally, to address data privacy concerns between units and the central server, the monitoring system aims to ensure that historical information, including the unit's observed feature vectors and rewards, remains confidential and distributed within each unit and is not exposed to the central server or other units. The online monitoring problem is further complicated under distributed data due to the need for efficient communication between local units and the central server and the increased uncertainty in model estimation. To solve this problem, this paper presents a federated collaborative online monitoring approach that integrates representation learning and a federated online learning algorithm. 

\subsection{Federated Collaborative Online Monitoring}
\begin{figure}[!t]
\centering
\includegraphics[width=0.8\textwidth]{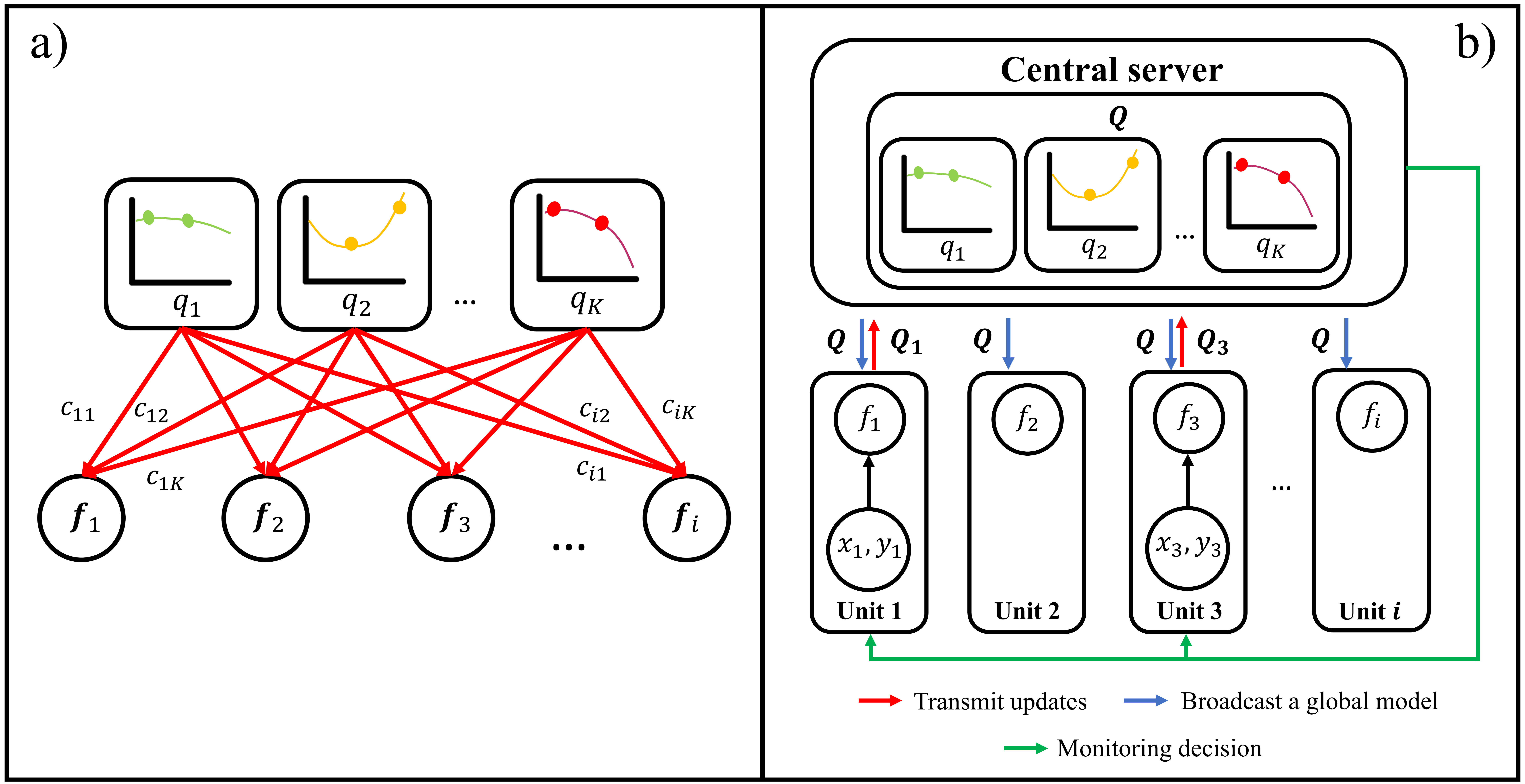}
\caption{a) Illustration of representation learning, b) The overall framework of the proposed Federated Collaborative Online Monitoring (FCOM) algorithm}
\label{fig:FigRep}
\end{figure}
\subsubsection{Representation learning-based modeling} \label{CL}
This paper assumes the reward function of each unit is linear with respect to the feature vectors i.e., $f_i(x_{it}) = \beta_i^T x_{it}$ where $\beta_i \in \mathbb{R}^{p}$ is the coefficient vector of unit $i$. As the units are related to each other, extracting the common structure between them could help their corresponding models learn more efficiently than considering each unit independently \cite{yang2021impact}. 
Following the representation learning approach introduced in \cite{yang2021impact}, $\beta_i$ is denoted as  $\beta_i = Qc_i$ where $Q = [q_1,\ldots, q_K] \in \mathbb{R}^{p \times K}$ is a low-rank matrix that represents $K$ representative models ($K \ll p$) in the reward functions. This representation captures the underlying group structure among units, a subset of which may follow the same representative model, as illustrated in Figure \ref{fig:FigRep}a). 
Each unit's coefficients on $K$ representative models are denoted as a membership vector $\{c_i\}_{i=1}^N, c_i \in \mathbb{R}^{K}$. Thus, the unit’s reward function can be illustrated as the weight combination of $K$ representative models $f_{i}(x_{it}) = \sum_k c_{ik} x_{it}^Tq_k $. 

In the monitoring problem, we aim to learn the representative models and units' membership vectors to make optimal decisions. Assuming parameters in $K$ representative models, $Q$, and units' membership vectors, $\{c_i\}_{i=1}^N$, are unconstrained, such a problem can be formulated as a global non-convex optimization problem in the following
\begin{equation}
     \min_{C, Q} \sum_{t} \sum_{i} \left\| x_{t, i}Qc_i - y_{t, i}\right\|_2^{2} +\eta_1\left\|Q\right\|_F^{2}
    +\eta_2\left\|C\right\|_F^{2}
     \label{eq:global_ob}
\end{equation}
where $C = [c_1,\ldots, c_N]$. The L2 regularization is used on $Q$ and $C$ to control the scale of unknown parameters with tuning parameters $\eta_1$ and $\eta_2$.

A few federated learning algorithms have been developed to solve the optimization problem similar to Equation \ref{eq:global_ob} from distributed data \cite{di2023ppfl,  jeong2022factorized}. Generally, they alternatively update the shared representation and membership vectors in each step. If the representation is fixed at $Q^*$, the problem of estimating the membership vectors from Equation \ref{eq:global_ob} can be achieved by decomposing it into a set of sub-problems that can be solved using local unit data. The feature matrix $\textbf{X}_{it} = [0,\ldots, x_{it}, \ldots, 0] \in \mathbb{R}^{p \times N}$ is transformed to $ X_{it} = \textbf{X}_{it} \otimes I_K \in \mathbb{R}^{Kp \times NK}$ where $I_K$ is the $K \times K$ identity matrix. Let $ \Tilde{c}_{it} = [0,\ldots, c_{it}, \ldots, 0] \in \mathbb{R}^{NK \times 1}, q = vec(Q) = [q_1^T,\cdots,q_K^T]^T \in \mathbb{R}^{Kp \times 1}$ where $vec(\cdot)$ is the vectorized operation, the sub-problem for each local unit $i$ is represented as follows. 
\begin{equation}
     \min_{c_{i}} \sum_{t=1}^{T} \left\| \Tilde{c}^T_{i} X^T_{it} q^* - y_{it} \right\|_2^{2}  + \eta_2\Vert c_{i} \Vert_2^2
     \label{eq:ob_c}
\end{equation}

If the membership vectors are fixed at $c_{i}^*$, the federated learning algorithms first estimate the representation in each local unit, denoted as $q_i$, and aggregate the local estimations in the central server to obtain an approximation of the global solution. The sub-problem for each local unit $i$ to estimate the representation is shown as follows.
\begin{equation}
     \min_{q_{i}} \sum_{t=1}^{T} \left\| \Tilde{c}^{*T}_{i} X^T_{it} q_i - y_{it} \right\|_2^{2}  + \eta_2\Vert q_{i} \Vert_2^2
     \label{eq:ob_q}
\end{equation}
However, these federated learning algorithms solve the representation learning problem in the offline setting, which is inappropriate for online monitoring where the models need to be updated with sequentially arrived data and the decisions need to be made under the exploitation-exploration trade-off. Thus, we propose a novel federated collaborative online monitoring algorithm to solve this issue in the next section.

\subsubsection{FCOM algorithm} \label{CL2}

We propose a federated collaborative online monitoring (FCOM) algorithm to estimate $q$ that is shared across $N$ units and $c_i$ which is specified for each unit from the distributed and sequentially observed data, as shown in Figure \ref{fig:FigRep}b). Assuming that a subset of units $\mathscr{A}_{t}$ is monitored in trial $t$ with new observations of monitoring rewards and feature vectors, the proposed method initially updates $c_i$ for monitored units and their local estimations of $q$, denoted as $q_i$, 
using an alternative least squares (ALS) algorithm that provides closed-form solutions as demonstrated in \textbf{Step-1}. Then, in \textbf{Step-2}, the updated estimations of $q_i$ in \textbf{Step-1} which are potential to benefit the global estimation are further uploaded to the central server. The central server utilizes these updates to estimate the global representation $q$ and broadcasts it to all units so that both monitored and not-monitored units can collaboratively update their models via the new observations. 

\textbf{Step-1 Local update:} There are two steps to estimate parameters in representation learning using an ALS algorithm to minimize 
Equation \ref{eq:ob_c} and \ref{eq:ob_q}. It firstly estimates $\Tilde{c}_{it}$ by fixing $q_{it}$ in \textbf{Step-1.1} and then estimates $q_{it}$ with fixed $\Tilde{c}_{it}$ in \textbf{Step-1.2}.

\textbf{Step-1.1:} When a local estimation of the representation $q_{i}$ is fixed, the estimation of a unit $i$'s membership vector $\Tilde{c}_{it}$ can be obtained by solving the objective function in Equation \ref{eq:ob_c}. The closed-form solution of $\Tilde{c}_{it}$ can be achieved by $\hat{\Tilde{c}}_{it} = D_{it}^{-1}d_{it}$ in which,
\begin{align}
    D_{it} ={}& \sum_{t^{\prime}=1}^t X_{it^{\prime}}^{T}q_{it}q_{it}^T X_{it^{\prime}} + \eta_2I_{NK}
    \label{eq:4} \\
    d_{it} ={}&  \sum_{t^{\prime}=1}^t X_{it^{\prime}}^{T}q_{it}y_{it^{\prime}}  
    \label{eq:5}
\end{align}

\textbf{Step-1.2:} When a unit $i$'s membership vector $\Tilde{c}_{i}$ is fixed, the estimation of the local representation $q_{it}$ can be obtained by solving the objective function in Equation \ref{eq:ob_q}. The closed-form solution of $q_{it}$ can be achieved by $\hat{q}_{it} = A_{it}^{-1}b_{it}$ in which,
\begin{align}
    A_{it} ={}&  \sum_{t^{\prime}=1}^t X_{it^{\prime}}\Tilde{c}_{it}\Tilde{c}_{it}^T X_{it^{\prime}}^{T} + \eta_1I_{Kp} \label{eq:6} \\
    b_{it} ={}& \sum_{t^{\prime}=1}^t X_{it^{\prime}}\Tilde{c}_{it}y_{it^{\prime}} \label{eq:7}
\end{align}

The unit $i$ alternatively solves these two steps until reaching convergence. The procedure of \textbf{Step-1} is summarized in Line 11 - Line 20 of \textbf{Algorithm \ref{alg:alg1}}. 

\textbf{Step-2 Global update:}  
The representation $q$ is shared across all $N$ units and needs to be estimated collaboratively. The update $\{\Delta A_{it}, \Delta b_{it}\}$ is defined as the information of $q$ that is learned from the local estimation of $q_{it}$ and denoted as
\begin{align}
    \Delta A_{it} ={}& \Delta A_{it-1} + \sum_{t^{\prime}=1}^t X_{it^{\prime}}\Tilde{c}_{it}\Tilde{c}_{it}^T X_{it^{\prime}}^{T} \label{eq:8} \\
    \Delta b_{it} ={}& \Delta b_{it-1} + \sum_{t^{\prime}=1}^t X_{it^{\prime}}\Tilde{c}_{it}y_{it^{\prime}} \label{eq:9} 
\end{align}

To avoid uploading new information every trial and reduce the communication cost, we introduce an event trigger for communication: only communicate/synchronize when local units have gathered sufficient new information.
The evaluation is in the form of the matrix determinant-based criterion, as shown in Equation \ref{eq:10} where $\gamma_{U_q} \geq 1$ is a hyperparameter:
\begin{align}
    \det(A_{it}) > \gamma_{U_q} \det(A_{it} - \Delta A_{it})
    \label{eq:10} 
\end{align}
If the unit $i$ passes Equation \ref{eq:10}, it uploads $\{\Delta A_{it}, \Delta b_{it}\}$ to the central server and reset $\Delta A_{it} = 0, \Delta b_{it} = 0$. The central server aggregates $\{\Delta A_{it}, \Delta b_{it}\}$ to its corresponding statistics $\{A_{gt}, b_{gt}\}$. After the aforementioned procedures are repeated to all monitored units, the central server estimates the global representation $\hat{q}_{t}$ as follows.
\begin{align}
    A_{gt} \mathrel{+}={}&  \Delta A_{it} \label{eq:11} \\
    b_{gt} \mathrel{+}={}&  \Delta b_{it} \label{eq:12} \\
    \hat{q}_{t} ={}& A_{gt}^{-1}b_{gt} \label{eq:q_g}
\end{align}
Finally, the central server broadcasts the representation $\hat{q}_{t}$ and its corresponding statistics to all units. Every unit $j \in [N]$ updates its local parameters $\hat{q}_{jt+1}$ and local statistics $\{A_{jt+1}, b_{jt+1}\}$ as follows.
\begin{align}
    A_{jt+1} ={}& A_{gt}\label{eq:14} \\
    b_{jt+1} ={}& b_{gt}\label{eq:15} \\
    \hat{q}_{jt+1} ={}& \hat{q}_{t}\label{eq:16}
\end{align}
The overall procedure of \textbf{Step-2} is summarized in Line 21 - Line 29 of \textbf{Algorithm \ref{alg:alg1}}.

\begin{algorithm}[!ht]
\caption{Federated collaborative
online monitoring (FCOM) algorithm}\label{alg:alg1}
\begin{algorithmic}[1]
\State \textbf{Input:} $\eta_1, \eta_2, \lambda, \alpha^q, \alpha^{\Tilde{c}}, M$
\State \textbf{Initialization:} units' statistics $A_{i0} \gets \eta_1 I_{Kp}$, $b_{i0} \gets 0$, $D_{i0} \gets \eta_2I_{NK}$, $d_{i0} \gets 0, \forall i \in [N]$; units' update $\Delta A_{i0} \gets 0$, $\Delta b_{i0} \gets 0$, $\Delta D_{i0} \gets 0$, $\Delta d_{i0} \gets 0, \forall i \in [N]$; the central server's statistics $A_{g0} \gets \eta_1 I_{Kp}$, $b_{g0} \gets 0$
\For{$t \gets 1,\ldots,T$}
    \For{all $i \in [N]$}
        \State Transform $ x_{it}$ to $X_{it} \in \mathbb{R}^{Kp \times NK}$
        \State \multiline{Compute $\hat{u}_{it} \gets \hat{\Tilde{c}}^T_{it} X_{it}^T \hat{q}_{it} + \alpha^{\Tilde{c}} \sqrt{q_{it}^T X_{it}D_{it}^{-1} X_{it}^{T}q_{it}}+ \alpha^q \sqrt{\Tilde{c}_{it}^T X_{it}^{T} A_{it}^{-1} X_{it}\Tilde{c}_{it}}$}
    \EndFor
    \State \multiline{The central server selects the subset $\mathscr{A}_{t}$ that includes $M$ units with highest UCB score computed in Line 6 }
    \State Each selected unit $i^{\prime}$ receives $y_{i^{\prime}t}$ from the environment 
    \For{all $i^{\prime} \in \mathscr{A}_{t}$}
        \While{not converge}
            \State $A_{i^{\prime}t} \gets A_{i^{\prime}t}  + X_{i^{\prime}t}\Tilde{c}_{i^{\prime}t}\Tilde{c}_{i^{\prime}t}^T X_{i^{\prime}t}^{T}, b_{i^{\prime}t} \gets b_{i^{\prime}t}  + X_{i^{\prime}t}\Tilde{c}_{i^{\prime}t}y_{i^{\prime}t}$
            \State $D_{i^{\prime}t} \gets D_{i^{\prime}t}  + X_{i^{\prime}t}^{T}q_{i^{\prime}t}q_{i^{\prime}t}^T X_{i^{\prime}t}, d_{i^{\prime}t} \gets d_{i^{\prime}t}  +  X_{i^{\prime}t}^{T}q_{i^{\prime}t}y_{i^{\prime}t}$ 

            \State $\Delta A_{i^{\prime}t} \gets \Delta A_{i^{\prime}t} + X_{i^{\prime}t}\Tilde{c}_{i^{\prime}t}\Tilde{c}_{i^{\prime}t}^T X_{i^{\prime}t}^{T}, \Delta b_{i^{\prime}t} \gets \Delta b_{i^{\prime}t} + X_{i^{\prime}t}\Tilde{c}_{i^{\prime}t}y_{i^{\prime}t}$ 

            \State $\hat{q}_{i^{\prime}t} \gets A_{i^{\prime}t}^{-1}b_{i^{\prime}t}, \hat{\Tilde{c}}_{i^{\prime}t} \gets D_{i^{\prime}t}^{-1}d_{i^{\prime}t}$
        \EndWhile
    
        \If {$\det(A_{i^{\prime}t}) > \gamma_{U_q} \det(A_{i^{\prime}t} - \Delta A_{i^{\prime}t})$}
            \State $A_{gt} \gets A_{gt} + \Delta A_{i^{\prime}t}$, $ b_{gt}  \gets b_{gt}  + \Delta A_{i^{\prime}t}, \hat{q}_{t} \gets A_{gt}^{-1}b_{gt}$ 
            \State Reset $\Delta A_{i^{\prime}t} \gets 0, \Delta b_{i^{\prime}t} \gets 0$
            
        \EndIf
    \EndFor
    
    \If {at least one unit triggers line 21}
        \State \multiline{The central server sends $\{A_{gt}, b_{gt}, \hat{q}_{t}\}$ to  all units $i \in [N]$.}
        \State \multiline{All units $i \in [N]$ update their local parameters $q_{it+1}$ and local statistics $\{A_{it+1}, b_{it+1}\}$.}
    \EndIf
\EndFor
\end{algorithmic}
\label{alg1}
\end{algorithm}

\subsubsection{Monitoring strategy design}
Based on the estimated representation and membership vectors, each unit can obtain a prediction of the monitoring reward in the next trial conditioning on its feature vector, i.e. $\hat{y}_{it+1} = \hat{c}^T_{it+1} X_{it+1}^T \hat{q}_{it+1}$. This prediction reflects the model's current best knowledge about the underlying group structure among all units and serves for exploitation purposes. Moreover, collecting only predicted rewards does not guarantee optimal selection, as it neglects prediction uncertainty estimated by each local unit. To address this issue, the upper confidence bound (UCB) principle is commonly used, which comprises two main components: the unit's predicted reward and the unit's prediction uncertainty \cite{auer2002using}. This score balances the objective of selecting units with the highest rewards (exploitation) and units with high uncertainty in reward predictions (exploration). This will allow the uncertain units to be explored to gain more information and confidence in selection. To obtain the UCB score, we need to derive the uncertainty of parameters estimated from the distributed data using the algorithm in Section \ref{CL2}. 

In the proposed model, the uncertainty of the estimated parameters is composed of two parts: the estimation uncertainty of the representation $\Vert\hat{q_{it}} - q^*\Vert_{A_{it}}$ and the estimation uncertainty of a unit's membership vector $\Vert\hat{\Tilde{c}}_{it} - \Tilde{c}^*\Vert_{D_{it}}$ where $q^*$ and $\Tilde{c}^*$ are the ground-truth parameters. Based on the closed-form solution in the ALS algorithm, the confidence ellipsoid of $\hat{q_{it}}$ and $\hat{\Tilde{c}}_{it}$ can be obtained by the \textbf{Lemma \ref{lemma1}} below.
\begin{lemma}
\label{lemma1}
When the Hessian matrices of the objective function in Equations \ref{eq:ob_c} and \ref{eq:ob_q} are positive definite at the optimizer $q^*$ and $\Tilde{c^*}$, for any $\epsilon_1 \geq 0$, $\epsilon_2 \geq 0$, $\Vert X_t\Vert_2 \leq S$, $\Vert q_{it}\Vert_2 \leq L$, $\Vert \Tilde{c}_{it}\Vert_2 \leq P$, and for any $\delta \geq 0$, with probability at least 1 - $\delta$, the estimation errors of the representation and unit's membership vector obtained from \textbf{Algorithm \ref{alg:alg1}} are upper bounded by:
\begin{align}
    \Vert\hat{q_t} - q^*\Vert_{A_{it}} &\leq \sqrt{Kp\ln(\frac{\eta_1 Kp + tS^2P^2}{\eta_1 Kp\delta})}  
    + \frac{2SPL}{\sqrt{\eta_1}}\frac{(v_1+\epsilon_1)(1-(v_1+\epsilon_1)^t)}{1-(v_1+\epsilon_1)} + \sqrt{\eta_1}L \label{eq:17}
\end{align}
\begin{align}
    \Vert \hat{\Tilde{c}}_{it} - \Tilde{c}^*\Vert_{D_{it}} 
    &\leq{} \sqrt{NK\ln(\frac{\eta_2NK + tS^2L^2}{\eta_2 NK\delta})} 
    + \frac{2SPL}{\sqrt{\eta_2}}\frac{(v_2+\epsilon_2)(1-(v_2+\epsilon_2)^t)}{1-(v_2+\epsilon_2)} + \sqrt{\eta_2} P \label{eq:18}
\end{align}
in which $0 < v_1 < 1$, $0 < v_2 < 1$ 
\end{lemma}
The detailed proof of \textbf{Lemma \ref{lemma1}} is provided in the Appendix. From \textbf{Lemma \ref{lemma1}}, we formulate \textbf{Proposition \ref{proposition1}} which defines the uncertainty of the predicted reward as follows. 
\begin{proposition}
\label{proposition1}
 With the probability at least $1-\delta$,
\begin{equation}
    \lvert y_{t,i}^* - \hat{y}_{t, i} \rvert \leq{}  \alpha^{\Tilde{c}} \sqrt{q_{it}^T X_{it}D_{it}^{-1} X_{it}^{T}q_{it}}+ \alpha^q \sqrt{\Tilde{c}_{it}^T X_{it}^{T} A_{it}^{-1} X_{it}\Tilde{c}_{it}} \label{eq:19}
\end{equation} 
\end{proposition}
where  $\alpha^q$ and $\alpha^{\Tilde{c}}$ are tuning parameters that are defined as the upper bounds of $\Vert\hat{q}_{it} - q^*\Vert_{A_{it}}$ and $\Vert\hat{\Tilde{c}}_{it} - \Tilde{c}^*\Vert_{D_{it}}$ respectively. The detailed proof of Equation \ref{eq:19} is shown in the Appendix. For the monitoring strategy, the central server selects $M$ units that have the highest UCB score. Denote $\alpha^q$ and $\alpha^{\Tilde{c}}$ as the upper bounds of $\Vert\hat{q}_{it} - q^*\Vert_{A_{it}}$ and $\Vert\hat{\Tilde{c}}_{it} - \Tilde{c}^*\Vert_{D_{it}}$ obtained from \textbf{Lemma \ref{lemma1}} respectively, the unit $i$'s UCB score which is based on the local statistics $\{A_{it}, b_{it}, D_{it}, d_{it}\}$ is defined as follow:
\begin{equation}
   \hat{u}_{it} = \hat{\Tilde{c}}^T_{it} X_{it}^T \hat{q}_{it} + \alpha^{\Tilde{c}} \sqrt{q_{it}^T X_{it}D_{it}^{-1} X_{it}^{T}q_{it}}+ \alpha^q \sqrt{\Tilde{c}_{it}^T X_{it}^{T} A_{it}^{-1} X_{it}\Tilde{c}_{it}} \label{eq:20}
\end{equation} 
Equation \ref{eq:20} illustrates the UCB score $\hat{u}_{it}$ which consists of two parts. The first term represents the predicted reward of unit $i$. The second and third terms represent the prediction uncertainty due to the unit's membership vector and representation estimations respectively. Since the resulting UCB score for each unit calculated from the estimated parameters is only exposed to the central server, it ensures privacy among all units as the central server cannot directly access the raw information (i.e., the unit's feature matrix and the unit's actual reward). 

The detailed procedure of the FCOM algorithm is summarized in \textbf{Algorithm \ref{alg:alg1}}. 

\subsection{Regret Analysis}
In this section, we provide a theoretical analysis of the cumulative regret of the proposed FCOM algorithm. Based on \textbf{Lemma \ref{lemma1}}, the alternating least squares-based parameter estimation satisfies the two inequalities shown in Equations \ref{eq:17} and \ref{eq:18} and it contributes to the final regret of the algorithm. \textbf{Theorem \ref{theorem1}} provides the upper regret bound and the communication cost of the FCOM algorithm, which analyzes the quality of the proposed selection strategy theoretically.

\begin{theorem}
\label{theorem1}
For any $\delta > 0$, with the probability at least 1 - $\delta$, the cumulative regret of the FCOM algorithm is upper bounded by:
\begin{align*}
    R(T) &\leq 2\alpha^{\Tilde{c}}\sqrt{2TNK\ln{(1 + \frac{TS^2L^2}{\eta_2 NK})}} 
    +  2\alpha^q G\sqrt{2TKp\ln{(1 + \frac{TS^2P^2}{\eta_1 Kp})}} 
    + \frac{2mv^2(1-v^{2T})}{1-v^2} 
\end{align*}
where $G = \min(\sqrt{N},\sqrt{[1+(N-1)(\gamma_{U_{q}}-1)]})$ and the communication cost $C(T) \leq \frac{N}{\ln{(\gamma_{U_q})}}Kp\ln(1+\frac{TS^2P^2}{\eta_1Kp})$. 

\end{theorem}
According to \textbf{Theorem \ref{theorem1}}, the upper regret bound can be decomposed into three components. The first term accounts for the regret associated with the estimation uncertainty of units' membership vectors $\Tilde{c}_{it}$. The second term reflects the regret stemming from the estimation uncertainty of shared representation $Q$, leveraging communication between units and the central server. The third term encompasses the regret due to q-linear convergence in estimating $\Tilde{c}_{it}$ and $Q$ over $T$ trials. In a regret comparison with the existing federated online learning algorithm that leverages mixed model for reward modeling (Sync-LinUCB) that yields an upper regret bound of $O(Np\sqrt{T}\ln{T})$ \cite{pmlr-v151-li22e}, the proposed FCOM algorithm achieves an improved regret bound of $O(NK\sqrt{T}\ln{T})$ by capturing the shared representation among units. This improvement is notable under the assumption that $K \ll p$. However, our algorithm incurs a higher communication cost of $O(Nkp\ln{T})$ compared to SynC-LinUCB, which is $O(Np\ln{T})$. The increase by a factor of $K$ is due to the number of global parameters estimated in the model increases. The proposed FCOM model estimates $K$ shared representative models, where the total number of global parameters is $Kp$. 
The Sync-LinUCB model, on the other hand, only estimates one global model with $p$ parameters.
\begin{figure*}[!t]
\centering
\includegraphics[width=\textwidth]{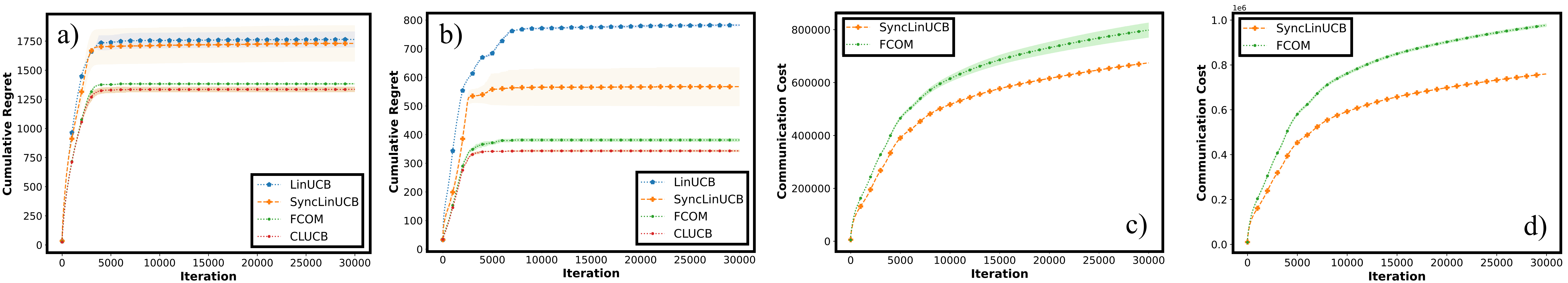} 
\caption{a)-b) The convergence of cumulative regret at $M = 33\%$ and $M = 66\%$,    c)-d) Communication cost at $M = 33\%$ and $M = 66\%$}
\label{fig:Fig2}
\end{figure*}
\begin{table*}[!t]
  \centering
  \caption{Cumulative regret comparison under different population size $N$}
  \label{tab:sa1}
  \begin{tabular}{@{}p{1.5cm}|ccccc@{}}
    \toprule
    & LinUCB & Sync-LinUCB & FCOM & CLUCB\\
    \midrule
    $N = 40$ & $1196.57 \pm 49.03$ & $ 1170.19 \pm 296.36 $ & $882.59 \pm 38.79$ & $812.72 \pm 8.75$\\
    $N = 80$ & $1605.92 \pm 32.36$ & $1203.05 \pm 15.08$ & $ 1117.53 \pm 10.78$ & $1061.15 \pm 17.29$\\
    $N = 100$ & $1764.24 \pm 68.64$ & $1730.04 \pm 155.83$ & $1383.90 \pm 6.90$ & $1335.70 \pm 22.89$\\
    $N = 120$ & $2141.11 \pm 79.36$ & $1808.56 \pm 181.11$ & $1553.23 \pm 56.99$ & $1378.40 \pm 70.72$\\
  \bottomrule
    \end{tabular}
\end{table*}

\section{Experiments} \label{Experiments}
\sloppy We evaluate the proposed method using both a simulation study and a real-world case study of cognitive degradation monitoring of Alzheimer's Disease (AD) in older populations. In both studies, we explore two monitoring scenarios to reflect varying levels of resource allocation in the real world. In the high-constraint scenario, only $33\%$ of the total population can be monitored in each monitoring cycle ($M=33\%$). In contrast, the relaxed-constraint scenario allows more units to be monitored, encompassing $66\%$ of the total population ($M=66\%$). We repeatedly monitor the populations over 30000 cycles ($T = 30000$) in all experiments.

The proposed FCOM algorithm is compared against various benchmark MAB algorithms, including LinUCB \cite{li2010contextual}, Sync-LinUCB \cite{pmlr-v151-li22e}, and CLUCB \cite{kosolwattana2023online}. LinUCB estimates the coefficients of units' reward functions independently through L2-regularized least square estimation. The Sync-LinUCB models personalized reward functions using a federated mixed model. Specifically, the random effects are estimated within local units independently whereas the fixed effect is estimated in the central server globally. However, neither algorithm exploits the representation learning approach to extract the latent patterns presented in the unit's reward function. CLUCB considers the online modeling and monitoring of units' reward functions via representation learning. However, it is developed for the centralized setting, meaning it shares and saves all units' observations of reward and feature vectors in the central server. The cumulative regret defined in Equation \ref{eq:2} is used to compare the performance of all algorithms in both simulation and real-world studies. The communication cost is also compared to illustrate the effectiveness of implementing a federated learning approach to solve the monitoring problem.  

\subsection{Simulation Studies}
We build the reward functions of $100$ units ($N=100$) and simulate their observations of monitoring reward over time, which is represented as $y_{it} = x_{it} \beta_i + \epsilon_{it}$. 
The feature vector $x_{it} = [x_{i1t},x_{i2t},\ldots,x_{ipt}]$ is simulated from $p$ sigmoid functions which are formulated by the following \cite{ venkatraghavan2019disease}.
\begin{equation}
    x_{ijt} = a_j + \frac{r_j}{1 + \text{exp}(-d_j(t-c_j))} + \epsilon_{it} \label{eq:21}
\end{equation}
The parameter $a_j, r_j, d_j, c_j$ and noise $\epsilon_{it}$ are randomly generated from a standard normal distribution. To guarantee the reward functions share a latent group structure, the coefficients in reward functions are assumed to be generated from $K$ representative models and a membership vector, denoted as $\beta_i = Qc_i$.
We consider three representative models ($K=3$) in this experiment.
The representative models are randomly generated from the multivariate normal distribution. To ensure each unit's reward function can be effectively resembled by one of the representative models, its membership vector is randomly generated from a mixture of Gaussian distributions. Specifically, the Gaussian distributions have zero-mean components, and the respective prior probabilities are estimated from the real data.:
\[ F_1(c) \sim N(0, \begin{bmatrix} \sigma^2 & 0 & 0\\ 0 & 1 & 0\\ 0 & 0 & 1 \end{bmatrix}),F_2(c) \sim N(0, \begin{bmatrix} 1 & 0 & 0\\ 0 & \sigma^2 & 0\\ 0 & 0 & 1 \end{bmatrix}),F_3(c) \sim N(0, \begin{bmatrix} 1 & 0 & 0\\ 0 & 1 & 0\\ 0 & 0 & \sigma^2 \end{bmatrix})\]
The covariance matrix in each distribution is a diagonal matrix with a $k^{th}$ diagonal element equal to $\sigma^2$ and the other elements equal to 1. $\sigma^2$ controls the significance of the latent group structure, where a larger value indicates more distinct groups. In this experiment, we set $\sigma^2$ at 100 to simulate three distinct groups. Random noises are generated from a standard normal distribution. 
We utilize k-means clustering to assign the initial value of $c_i$ based on the centroids learned from the similarity matrix where each element in the matrix is calculated by the cosine similarity of the membership vectors. We then use $c_i$ to initialize the value of $Q$ by following the update part from lines 12, and 15 in \textbf{Algorithm \ref{alg:alg1}}. The experiment is repeated 3 times to estimate the variation in performance.

In Figure \ref{fig:Fig2} a) and b), the cumulative regret of various bandit algorithms is illustrated under two monitoring scenarios. Sync-LinUCB exhibits lower cumulative regret than LinUCB in both scenarios but it has high variations in the high-constraint scenario. 
Our proposed FCOM algorithm achieves a lower cumulative regret than LinUCB and Sync-LinUCB, despite having slightly higher communication costs than Sync-LinUCB as depicted in Figure \ref{fig:Fig2} c) and d). This aligns with the theoretical analysis outlined in the previous section. The results indicate that incorporating the shared representation and providing information about the group structure among units, leads to the most significant performance improvement. Moreover, compared to CLUCB which is in a centralized setting, the cumulative regret of FCOM is slightly higher. This suggests modeling the dynamic health progression in a federated learning setting to preserve data privacy does not hurt the performance of the proposed algorithm.

We further conduct sensitivity analysis on different parameters, including the population size ($N$) and the number of representative models ($K$), to investigate the sensitivity of model performance, as illustrated in Tables \ref{tab:sa1} and \ref{tab:sa2}. Table \ref{tab:sa1} presents the cumulative regret of four algorithms under varying population sizes. It demonstrates that the cumulative regret of LinUCB increases linearly as $N$ increases. Although Sync-LinUCB achieves regret reduction compared to LinUCB, it exhibits high variations across multiple population sizes. In contrast, the cumulative regret of FCOM is lower than the aforementioned benchmark algorithms and does not increase linearly with an increase in $N$. While its cumulative regret is slightly higher than CLUCB, it displays low variations compared to LinUCB and Sync-LinUCB. This indicates the robustness of our proposed algorithm against different population sizes. Table \ref{tab:sa2} displays the cumulative regret under different numbers of representative models ($K$). The cumulative regrets of all algorithms increase as $K$ increases, reflecting the increased difficulty of the monitoring problem under a more complex latent structure. However, the cumulative regret of our proposed method increases marginally compared to LinUCB and Sync-LinUCB, suggesting the stability of our proposed algorithm as the group structure becomes more complex.
\begin{table*}[!t]
\centering
  \caption{Cumulative regret comparison under different number of representative models $K$}
  \label{tab:sa2}
  \begin{tabular}{@{}p{1cm}|ccccc@{}}
    \toprule
    & LinUCB & Sync-LinUCB & FCOM & CLUCB\\
    \midrule
    $K = 3$ & $1764.24 \pm 68.64$ & $1730.04 \pm 155.83$ & $1383.90 \pm 6.90$ & $1335.70 \pm 22.89$\\
    $K = 5$ & $ 2246.64\pm 170.36$ & $ 2166.59\pm 82.38$ & $1491.58 \pm 25.78$ & $1484.91 \pm 26.71$\\
  \bottomrule
\end{tabular}
\end{table*}
\subsection{Application to online cognitive monitoring in Alzheimer's disease (AD)}

A case study of applying the proposed method to online monitor the cognitive degradation in older adults is further conducted. Monitoring the cognitive degradation in older adults is critical for the early detection of Alzheimer's disease (AD). The monitoring reward is defined as the cognitive status measured by the Mini-Mental State Examination (MMSE) \cite{MUELLER2005869} in each cycle. The MMSE score is based on the severity of degradation which is categorized into three states including healthy state (27-30), mild cognitive impairment (24-26), and dementia (0-23) \cite{MUELLER2005869}. A more severe cognitive status corresponds to a higher monitoring reward as it can lead to the early prevention and treatment of AD. Due to the limited monitoring resources in clinical practice, the objective of this case study is to apply the proposed method to maximally monitor older adults with severe cognitive status under limited resources. To apply the proposed method, we model the progression of MMSE by the commonly used degradation model that uses time and its polynomial basis functions
as predictors in which the cognitive degradation process of the patient $i$ is formulated as \cite{ bartzokis2004heterogeneous}:
\begin{equation} \label{eq:ex2}
y_{it} = \beta_{i0} + \beta_{i1}t + \beta_{i2}t^2 + \beta_{i3}t^3 + \beta_{i4}t^4 + \beta_{i5}t^5 +\epsilon_{it}
\end{equation}
where $y_{it}$ is the MMSE measurement of the patient $i$ at time point $t$. The dataset is acquired from Alzheimer's Disease Neuroimaging Initiative (ADNI) \cite{MUELLER2005869} which consists of longitudinal measurements of MMSE for 648 older adults. We repeatedly sample 100 older adults from the dataset 5 times to estimate the variation in performance. They are collected at baseline, $12^{th}, 24^{th}, 36^{th}, 48^{th}$, and $60^{th}$ month. The measurements collected from baseline to the $60^{th}$ month are equally divided into 30,000 cycles. We use linear interpolation to impute the missing MMSE values for each older adult. 
\begin{figure}[!t]
\centering
\includegraphics[width=0.7\columnwidth]{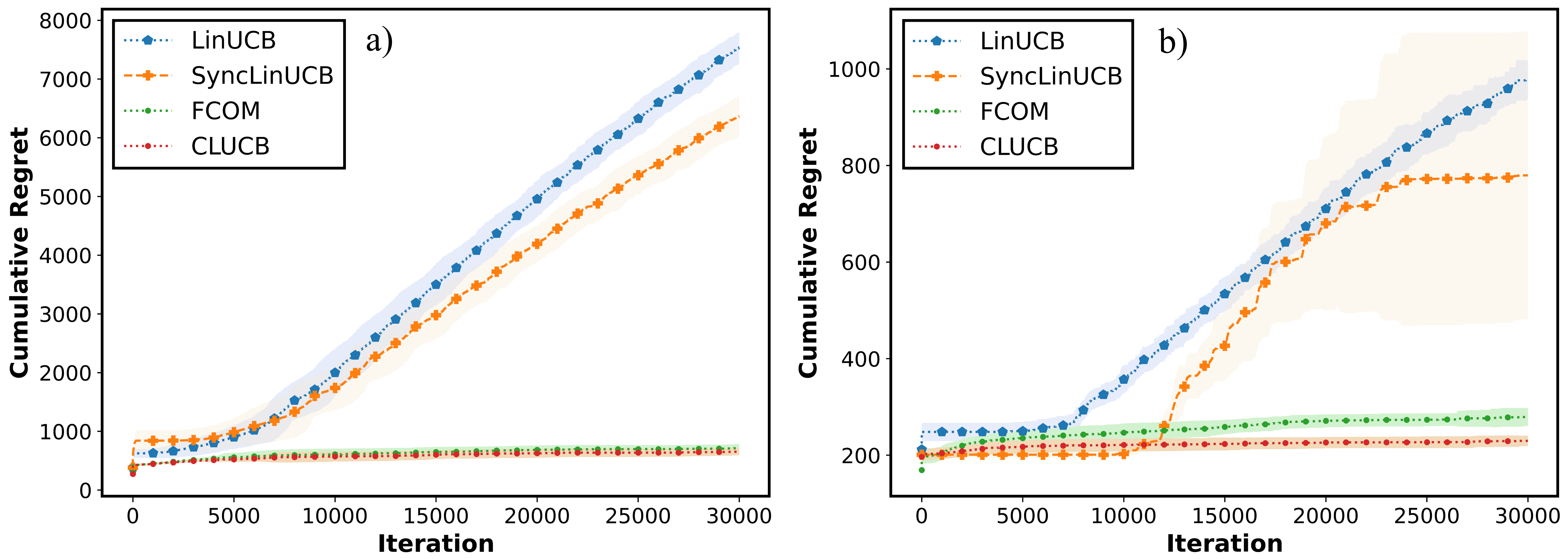}
\caption{The cumulative regret for all bandit algorithms shown in the Alzheimer’s Disease dataset with a) $M = 33\%$ and b) $M = 66\%$}
\label{fig:fig3}
\end{figure}

The cumulative regrets of different algorithms under two monitoring scenarios are compared in Figure \ref{fig:fig3}. LinUCB and Sync-LinUCB exhibit higher cumulative regret than the proposed FCOM in both scenarios. The cumulative regret of Sync-LinUCB reaches convergence only in the high-constraint scenario, while our proposed algorithm converges and outperforms Sync-LinUCB in both scenarios. This observation suggests that effectively leveraging representation learning improves the accuracy of cognitive degradation modeling in AD and enables more accurate monitoring resource allocation. Moreover, it indicates the stability of the proposed algorithm regardless of the number of selected older adults. Moreover, similar to the simulation studies, the cumulative regret of the proposed FCOM is slightly higher than CLUCB, suggesting that modeling older adults' cognitive degradation in a decentralized setting to preserve data privacy does not hurt the performance of the proposed algorithm.

\section{Conclusions \& future work} \label{Conclusion}

This paper presents a novel Federated Online Collaborative Monitoring (FCOM) algorithm designed to enable decentralized online learning of a population comprising latent structured units. The proposed FCOM algorithm leverages representation learning to capture the latent group structure and estimates the latent group structure from distributed and sequentially observed data by real-time updating the local statistics and communicating these statistics between the central server and local units. A novel UCB score is further developed to inform the optimal monitoring resource allocation. Through regret analysis, it is demonstrated that the proposed algorithm yields a reduced upper regret bound compared to the benchmark models when the latent group structure is low-rank. Simulation studies and an empirical study of online cognitive monitoring in Alzheimer’s disease further validate that the proposed algorithm achieves lower cumulative regret compared to other benchmark models. In the future, we will extend the proposed FCOM algorithm to an asynchronous setting where not all units are available at the beginning of the trial. Moreover, we would like to develop more efficient communication by minimizing the size of statistics to eliminate the necessity of incurring a higher communication cost while still achieving a lower regret bound. Lastly, we will integrate shared similarity information between units to facilitate the inductive transfer of knowledge and preserve privacy.


\addtolength{\textheight}{-5cm} 
\bibliographystyle{IEEEtran}
\bibliography{References}

\end{document}